\documentclass{article}

\usepackage[final]{nips_2017}

\usepackage[utf8]{inputenc} % allow utf-8 input
\usepackage[T1]{fontenc}    % use 8-bit T1 fonts
\usepackage{hyperref}       % hyperlinks
\usepackage{url}            % simple URL typesetting
\usepackage{booktabs}       % professional-quality tables
\usepackage{amsfonts}       % blackboard math symbols
\usepackage{nicefrac}       % compact symbols for 1/2, etc.
\usepackage{microtype}      % microtypography
\usepackage{amsmath,amssymb,amsopn,amsthm,bm,bbm}
\usepackage{multirow,rotating}
\usepackage{todonotes}

\newtheorem{thm}{Theorem} 

\newtheorem{definition}{Definition}

\newcommand{\Y}{\mathcal{Y}}
\newcommand{\X}{\mathcal{X}}
\newcommand{\E}{\mathbb{E}}

\newcommand{\D}{\mathcal{D}}

\newcommand{\cDC}{$\hat{c}$\,DC}
\newcommand{\cDR}{$\hat{c}$\,DR}

\title{Counterfactual Learning for Machine Translation: Degeneracies and Solutions}

\author{Carolin Lawrence \\
	Computational Linguistics,\\Heidelberg University,\\Germany \\\texttt{lawrence@cl.uni-heidelberg.de}\\\And
	Pratik Gajane \\
	INRIA SequeL / Orange labs,\\ France\\
	\texttt{pratik.gajane@gmail.com} \\\And
	Stefan Riezler\\
	Computational Linguistics \& IWR,\\Heidelberg University,\\Germany\\\texttt{riezler@cl.uni-heidelberg.de}}

\begin{document}
	
	\maketitle
	
	\begin{abstract}
		Counterfactual learning is a natural scenario to improve web-based machine translation services by offline learning from feedback logged during user interactions. In order to avoid the risk of showing inferior translations to users, in such scenarios mostly exploration-free deterministic logging policies are in place. We analyze possible degeneracies of inverse and reweighted propensity scoring estimators, in stochastic and deterministic settings, and relate them to recently proposed techniques for counterfactual learning under deterministic logging. 
	\end{abstract}

	\section{Introduction}

Machine translation has recently become a commodity service that is offered for free for online translation, e.g, by Google or Microsoft, or is integrated into e-commerce platforms (eBay) or social media (Facebook). Such commercial settings facilitate the collection of user feedback on the quality of machine translation output, either in form of an explicit user rating, or as indirect signal that can be inferred from the interaction of the user with the translated content. While user feedback in form of user clicks on displayed ads has been shown to be a valuable signal in response prediction for online advertising \citep{ChapelleLi:11,BottouETAL:13}, the gold mine of free user feedback has not yet been exploited in the area of machine translation. Recent research has proposed bandit structured prediction \citep{SokolovETALnips:16,KreutzerETAL:17,NguyenETAL:17} for online learning of machine translation from weak user feedback to predicted translations, instead of from costly manually created reference translations. The scenario investigated in these works is still far removed from real world applications in commercial machine translation: Besides the fact that previous research has been confined to simulated feedback, online bandit learning is unrealistic in commercial settings due to the additional latency and the desire for offline testing of system updates before deployment. A natural solution would be to exploit counterfactual learning that reuses existing interaction data where the predictions have been made by a historic system different from the target system. However, both online learning and offline learning from logged data are plagued by the problem that exploration is prohibitive in commercial systems since it means to show inferior translations to users. This effectively results in deterministic logging policies that lack explicit exploration, making an application of off-policy methods theoretically questionable.

\cite{LawrenceETAL:17} recently showed that bandit learning of machine translation is possible even under deterministic logging. They proposed an application of techniques such as doubly-robust policy evaluation and learning \citep{DudikETAL:11} or weighted importance sampling \citep{JiangLi:16,ThomasBrunskill:16} to offline learning from deterministically logged data, and presented evidence from simulation experiments that confirmed their conjecture that these techniques effectively serve to smooth out deterministic components. The purpose of our paper is to give a formal account on the possible degeneracies of the standard inverse propensity scoring technique \citep{RosenbaumRubin:83} and its reweighted variant \citep{Kong:92} under stochastic and deterministic logging, with the goal of a clearer understanding of the effectiveness of the techniques proposed in \cite{LawrenceETAL:17}.

	\section{Counterfactual Learning for Machine Translation}

In the following, we give a short overview of the methods developed in \cite{LawrenceETAL:17}. They formalize the problem of counterfactual learning  for bandit structured prediction as follows: $\X$ denotes a structured input space, $\Y(x)$ denotes the set of possible structured output for input $x$, and $\delta: \Y  \rightarrow [0,1]$ denotes a reward function quantifying the quality of structured outputs. A data log is denoted as a set of tuples $\D = \{(x_t,y_t,\delta_t)\}_{t=1}^n$, where for inputs $x_t$, a logging policy $\mu$ produced an output $y_t$, which is logged with a corresponding reward $\delta_t \in [0,1]$. In the case of stochastic logging, a propensity score $\mu(y_t|x_t) \in (0,1]$ is logged in addition. Using the inverse propensity scoring approach (IPS), importance sampling achieves an unbiased estimate of the expected reward under the parametric target policy $\pi_w(y_t|x_t) \in [0,1]$:
\begin{align}
\label{eq:ips}
\hat{V}_{\text{IPS}}(\pi_w) & = \frac{1}{n} \sum_{t=1}^n \delta_t \frac{\pi_w(y_t|x_t)}{\mu(y_t|x_t)} \\ \notag
& \approx \E_{p(x)} \E_{\mu(y|x)} [\delta(y) \frac{\pi_w(y|x)}{\mu(y|x)}] \\ \notag
& =  \E_{p(x)} \E_{\pi_w(y|x)} [\delta(y)]. \\ \notag
\end{align}
In case of deterministic logging, outputs are logged with propensity $\mu_t =1, t=1, \ldots, n$ of the historical system. This results in empirical risk minimization, or empirical reward maximization, without correction of the sampling bias of the logging policy:
\begin{align}
\label{eq:emp-risk} 
\hat{V}_{\text{DPM}}(\pi_w) = \frac{1}{n} \sum_{t=1}^n \delta_t \pi_w(y_t|x_t).
\end{align}
\cite{LawrenceETAL:17} call equation \eqref{eq:emp-risk} the \emph{deterministic propensity matching (DPM)} objective, and propose a first modification by the use of weighted importance sampling \citep{PrecupETAL:00,JiangLi:16,ThomasBrunskill:16}. The new objective is the \emph{reweighted deterministic propensity matching (DPM+R)} objective:
\begin{align}
\label{eq:r} 
\hat{V}_{\text{DPM+R}}(\pi_w) &= \frac{1}{n}\sum_{t=1}^{n} \delta_t \bar{\rho}_w(y_t|x_t) \\ \notag
&= \frac{\frac{1}{n}\sum_{t=1}^{n} \delta_t \rho_w(y_t|x_t)}{\frac{1}{n}\sum_{t=1}^{n} \rho_w(y_t|x_t)},
\end{align}
with $\rho_w(y_t|x_t) = \pi_w(y_t|x_t)$. Setting $\rho_w(y_t|x_t) = \frac{\pi_w(y_t|x_t)}{\mu(y_t|x_t)}$ recovers IPS with reweighting, $\hat{V}_{\text{IPS+R}}(\pi_w)$ \citep{SwaminathanJoachimsNIPS:15}.

\cite{LawrenceETAL:17} present further modifications of Equation \eqref{eq:r} by the incorporation of a direct reward estimation method into IPS as proposed in the doubly-robust (DR) estimator \citep{DudikETAL:11,JiangLi:16,ThomasBrunskill:16}. Let $\hat{\delta}(x_t, y_t)$ be a regression-based reward model trained on the logged data, and let $\hat{c}$ be a scalar that allows to optimize the estimator for minimal variance \citep{Ross:13}. They define a \emph{doubly controlled} empirical risk minimization objective $\hat{V}_{\text{\cDC}}$ as follows:
\begin{align}
\label{eq:dr-risk} 
\hat{V}_{\text{\cDC}}(\pi_w) =& \frac{1}{n} \sum_{t=1}^{n} \Big[  (\delta_t-\hat{c}\hat{\delta_t}) \; \bar{\rho}_w(y_t|x_t) +  \hat{c} \sum_{y \in \mathcal{Y}(x_t)} \hat{\delta}(x_t,y) \; \rho_w(y | x_t) \Big], 
\end{align}
with $\rho_w(y_t|x_t) = \pi_w(y_t|x_t)$. Setting $\hat{c} = 1$ yields an objective called $\hat{V}_{\text{DC}}$. Setting $\rho_w(y_t|x_t) = \frac{\pi_w(y_t|x_t)}{\mu(y_t|x_t)}$ recovers the standard stochastic doubly-robust estimator $\hat{V}_{\text{\cDR}}$. The optimal scalar parameter $\hat{c}$ can be derived easily by taking the derivative of the variance term, leading to $\hat{c} = \frac{\text{Cov}(X,Y)}{\text{Var}(Y)}.$

\begin{table*}[t]
	\begin{center}
		\caption{Gradients of counterfactual objectives.}
		\label{tab:gradients}
		\begin{tabular}{l}
			\toprule
			$\nabla \hat{V}_{\text{IPS/DPM}} =  \frac{1}{n} \sum_{t=1}^{n} \delta_t \rho_w(y_t | x_t) \nabla \log \pi_w(y_t | x_t).$ \\ \midrule
			$\nabla\hat{V}_{\text{IPS+R/DPM+R}} = \frac{1}{n} \sum_{t=1}^{n} [ \delta_t \bar{\rho}_w(y_t|x_t) (  \nabla \log \pi_w(y_t | x_t)  - \sum_{u=1}^{n} \bar{\rho}_w(y_u|x_u)  \nabla \log \pi_w(y_u | x_u) )].$ \\ \midrule
			$\nabla\hat{V}_{\text{\cDC/\cDR}} = \frac{1}{n} \sum_{t=1}^{n} [ (\delta_t-\hat{c}\hat{\delta}) \bar{\rho}_w(y_t|x_t) (  \nabla \log \pi_w(y_t | x_t) - \sum_{u=1}^{n} \bar{\rho}_w(y_u|x_u)  \nabla \log \pi_w(y_u | x_u)) $ \\
			$ \qquad \qquad \qquad + \hat{c} \sum_{y \in \mathcal{Y}(x_t)} \hat{\delta}(x_t,y)  \pi_w(y | x_t) \nabla \log \pi_w(y | x_t)].$ \\
			\bottomrule
		\end{tabular}
	\end{center}
\end{table*}

The learning algorithms in \cite{LawrenceETAL:17} are defined by applying a stochastic gradient ascent update rule $w_{t+1} = w_t + \eta \nabla\hat{V}(\pi_w)_t$ to the objective functions defined above. The gradients are shown in Table \ref{tab:gradients}. In the experiments reported in \cite{LawrenceETAL:17}, the policy distribution is assumed to be a  Gibbs model
$
\pi_w(y_t | x_t) = \frac{e^{\alpha (w^{\top} \phi(x_t,y_t))}}{\sum_{y\in \mathcal{Y}(x_t)} e^{\alpha (w^{\top} \phi(x_t, y))}},
$
based on a feature representation $\phi:\X \times \Y \rightarrow \mathbb{R}^d$, a weight vector $w \in \mathbb{R}^d$, and a smoothing parameter $\alpha \in \mathbb{R}^{+}$, yielding the following simple derivative $\nabla \log \pi_w(y_t | x_t) = \alpha \big(\phi(x_t,y_t) - \sum_{y\in \mathcal{Y}(x_t)}  \phi(x_t, y)\pi_w(y_t | x_t)\big).$

	\section{Degenerate Behaviour in Counterfactual Learning}\label{sec:degenerate}

Both the IPS and the DPM estimators can exhibit a degenerate behavior in that they can be maximized by simply setting all logged outputs to probability $1$, i.e.,  if $\pi_w(y_t|x_t) = 1$ for $\forall (y_t, x_t, \delta_t) \in \D = \{(x_t,y_t,\delta_t)\}_{t=1}^n$. This is the case irrespective of whether data are logged stochastically (IPS) or deterministically (DPM). Obviously, this is undesired as the probability for low reward outputs should not be raised.  For abbreviation, we set $\pi_t = \pi_w(y_t | x_t)$ and $\mu_t = \mu (y_t | x_t)$.

\begin{thm}\label{thm:1}
	$max_{\pi} \hat{V}_{\text{IPS}} \land max_{\pi} \hat{V}_{\text{DPM}} \textrm{ if } \forall (y_t, x_t, \delta_t) \in \D : \pi(y_t|x_t) = 1 \land \delta_t > 0$.
\end{thm}

\textit{Proof.} We start by showing that the value of $\hat{V}_{\text{IPS}}$ where $\forall (y_t, x_t, \delta_t) \in \D: \pi_t = 1$ is greater than the value of $\hat{V}_{\text{IPS}}$ where $\exists (x_t,y_t,\delta_t):\pi_t \in [0,1)$.
 W.l.og. assume that $(x_n, y_n, \delta_n)$ is the tuple with $\pi_t \in [0,1)$. Then
\begin{align}
\sum_{t=1}^{n} \frac{\delta_t}{\mu_t} 
&> \sum_{t=1}^{n-1} \frac{\delta_t}{\mu_t} + \frac{\delta_n \pi_n}{\mu_n},\\ \notag
\frac{\delta_n}{\mu_n}
&>\frac{\delta_n \pi_n}{\mu_n},\\ \notag
1
&>\pi_n,
\end{align}

where the last line is true by assumption $\pi_n \in [0,1)$. Because DPM is a special case of IPS with $\mu_t = 1$ for $\forall (y_t, x_t) \in \D = \{(x_t,y_t,\delta_t)\}_{t=1}^n$, the proof also holds for DPM. $\qed$

The degenerate behavior of IPS and DPM described in Theorem \ref{thm:1} can be fixed by using reweighting, which results in defining a probability distribution over the log $\D$. Under reweighting, increasing the probability of a low reward output takes away probability mass from the higher reward output. This decreases the value of the estimator, and will thus be avoided in learning.

However, IPS+R and DPM+R still can behave in a degenerate manner, as we will show in the following. We define the set $\D^{max}$ that contains all tuples that receive the highest reward $\delta_{max}$ observed in the log, and we assume $\delta_{max} > 0$, leading to a cardinality of $\D^{max}$ of at least one.
%\todo{Stefan: Removed footnote introducing $\delta_{max}$. Please check!}

\begin{definition}\label{def:two_sets}
  Let $\D^{max} = max_{\delta} \D$, then $\D = \D^{max} \cup \D \backslash \D^{max}$.
\end{definition}

We will show that the estimators can be maximized by simply setting the probability of at least one tuple in $\D^{max}$ to a value higher than $0$, while leaving all other tuples in $\D^{max}$ at their probabilities  $[0,1]$, and setting the  probability of tuples in the set $ \D \backslash \D^{max}$ to $0$. Clearly, this is undesired as outputs with a reward close to $\delta_{max}$ should not receive a probability of $0$. Furthermore, this learning goal is easy to achieve since a degenerate estimator only needs to be concerned about lowering the probability of tuples in $\D \backslash \D^{max}$ as long as there is one tuple of $\D^{max}$ with a probability above 0. We want to prove the following theorem:
%\todo{Stefan: Small change in description. Please check!}

\begin{thm}\label{thm:reweight}
$max_{\pi} \hat{V}_{\text{IPS+R}} \land max_{\pi} \hat{V}_{\text{DPM+R}} \textrm{ if } \exists (x_t, y_t, \delta_{max}) \in \D^{max} : \pi_t \in (0,1] \land \forall (y_t, x_t, \delta_t) \in \D \backslash \D^{max} : \pi_t = 0$.
\end{thm}

We introduce a definition of data indices belonging to the sets $\D^{max}$ and its complement in $\D$:

\begin{definition}
	Let \[\hat{V}_{\text{IPS+R}}(\pi_w) = \frac{\sum_{t=1}^{n} \delta_t \frac{\pi_t}{\mu_t}}{\sum_{t=1}^{n} \frac{\pi_t}{\mu_t}} 
	= \frac{\delta_{max} \sum_{t=1}^{s-1} \frac{\pi_t}{\mu_t} +   \sum_{t=s}^{n} \delta_t  \frac{\pi_t}{\mu_t} }{\sum_{t=1}^{n} \frac{\pi_t}{\mu_t}},\]
	
	where w.l.o.g. indices $(1 \dots (s-1))$ refer to tuples in $\D^{max}$ and indices $(s \dots n)$ refer to indices in $\D \backslash \D^{max}$. Thus, $\D^{max} = \{(x_t,y_t,\delta_{max})\}_{t=1}^{s-1}$ and $\D \backslash \D^{max} = \{(x_t,y_t,\delta_t)\}_{t=s}^n$.
\end{definition}

\textit{Proof.} We need to show that the value of $\hat{V}_{\text{IPS+R}}$ where $\pi_t = 0$ for $\forall (y_t, x_t, \delta_t) \in\D^{max}$ is lower than the value of $\hat{V}_{\text{IPS+R}}$ where $\exists (x_t, y_t, \delta_{max}) \in \D^{max}$ with $\pi_t \in (0,1]$. Then
%\end{prop}

\begin{align}
\frac{\sum_{t=s}^{n} \delta_t \frac{\pi_t}{\mu_t} }{\sum_{t=s}^{n} \frac{\pi_t}{\mu_t}}&
<
\frac{\delta_{max} \sum_{t=1}^{s-1} \frac{\pi_t}{\mu_t} +  \sum_{t=s}^{n} \delta_t \frac{\pi_t}{\mu_t} }{\sum_{t=1}^{n} \frac{\pi_t}{\mu_t}}\\ \notag
0&<\frac{\delta_{max} \sum_{t=1}^{s-1} \frac{\pi_t}{\mu_t} }{\sum_{t=1}^{n} \frac{\pi_t}{\mu_t}},\notag
\end{align}
where the last line is true for $\delta_{max}>0$ as long as $\exists (x_t, y_t, \delta_{max}) \in \D^{max}$ with $\pi_t > 0$ as $\mu_t \in (0,1]$ by definition.

Furthermore, we need to show that the value of $\hat{V}_{\text{IPS+R}}$ where $\exists (y_t, x_t, \delta_t) \in \D \backslash \D^{max}$ with $\pi_t \in (0,1]$  is lower than the value of $\hat{V}_{\text{IPS+R}}$ with $\pi_t = 0$ for $\forall (y_t, x_t, \delta_t) \in \D \backslash \D^{max}$.

From the above, it is clear that $\exists (x_t, y_t, \delta_{max}) \in \D^{max}$ with $\pi_t \in (0,1]$, thus $\frac{\delta_{max} \sum_{t=1}^{s-1} \frac{\pi_t}{\mu_t} }{\sum_{t=1}^{s-1} \frac{\pi_t}{\mu_t}}$ is defined.
W.l.o.g. assume that $(y_s, x_s, \delta_s) \in \D \backslash \D^{max}$ is the tuple with $\pi_s \in (0,1]$. Then 

\begin{align}\label{eq:to_proof_th2}
\frac{\delta_{max} \sum_{t=1}^{s-1} \frac{\pi_t}{\mu_t} +  \delta_s \frac{\pi_s}{\mu_s} + \sum_{t=s+1}^{n}\delta_t\frac{0}{\mu_t}}{\sum_{t=1}^{s} \frac{\pi_t}{\mu_t}+\sum_{t=s+1}^{n}\frac{0}{\mu_t}}
&<
\frac{\delta_{max} \sum_{t=1}^{s-1} \frac{\pi_t}{\mu_t} + \sum_{t=s}^{n} \delta_t \frac{0}{\mu_t}}{\sum_{t=1}^{s-1} \frac{\pi_t}{\mu_t} + \sum_{t=s}^{n} \frac{0}{\mu_t}} =  \delta_{max},\\ \notag
\delta_{max} \sum_{t=1}^{s-1} \frac{\pi_t}{\mu_t} 
+ \delta_s \frac{\pi_s}{\mu_s} 
&<
\delta_{max}\sum_{t=1}^{s} \frac{\pi_t}{\mu_t},\\ \notag
\delta_{max} \sum_{t=1}^{s-1} \frac{\pi_t}{\mu_t} 
+ \delta_s \frac{\pi_s}{\mu_s} 
&<
\delta_{max}\sum_{t=1}^{s-1} \frac{\pi_t}{\mu_t}
+ \delta_{max} \frac{\pi_s}{\mu_s},\\ 
\delta_s \frac{\pi_s}{\mu_s} 
&<
\delta_{max} \frac{\pi_s}{\mu_s}.\label{eq:ipsr_proof}
\end{align}

Equation \ref{eq:ipsr_proof} is true as $\delta_s < \delta_{max}$ by Definition \ref{def:two_sets} . 
As DPM+R is a special case of IPS+R with $\mu_t = 1$ for $\forall (y_t, x_t) \in \D = \{(x_t,y_t,\delta_t)\}_{t=1}^n$, the proof also holds for DPM+R. $\qed$

While employing stochastic gradient ascent, IPS+R and DPM+R can be prevented from reaching their degenerate state by performing early stopping on a validation set. However, one cannot control what happens to the probability mass that is freed when lowering the probability of a logged output. The freed probability mass could be allocated to outputs that receive a lower reward than the logged output which would create a system that is worse than the logging system.

The estimators $(\hat{c})$DR and $(\hat{c})$DC successfully solve this problem. The direct reward predictor takes the whole output space into account and thus assigns rewards to any structured output. The objective will now be increased if the probability of outputs with high estimated reward is increased, and decreased for outputs with low estimated reward. For this to happen, high reward outputs other than the ones with maximal reward will be considered, even if the outputs have not been seen in the training log. This will shift probability mass to unseen data with high estimated reward, which is a desired property in learning.

	\section{Experimental Evidence}
           
For completeness, we report the experimental evidence that \cite{LawrenceETAL:17} provide to show the effectiveness of their proposed techniques. They report an application of counterfactual learning in a domain-adaptation setup for machine translation. A model is trained using out-of-domain data using the hierarchical phrase-based machine translation framework that is based on a linear learner. The model is given in-domain data to translate, and outputs are logged together with their per-sentence BLEU score to the true reference, which simulates the reward signal. Experiments are conducted on two language pairs. The first is German-to-English and its baseline system is trained on the concatenation of the Europarl corpus, the Common Crawl corpus and the News corpus. The target domain is represented by a corpus containing transcribed TED talks. The second language pair is French-to-English. Its out-of-domain system is trained on the Europarl corpus and the target domain is the News corpus.

\begin{table*}[!htbp]
	\centering
	\caption{BLEU increase over the out-of-domain baseline on validation and test set for deterministically and stochastically created logs.} 
	\label{exp:learn_results}
	\begin{tabular}{lllccccc}
		&&&BLEU&\multicolumn{3}{c}{BLEU difference}&BLEU\\
		&&&out-of-domain&DPM+R&DC&\cDC{}&in-domain\\
		\toprule
		\multirow{4}{*}{\begin{sideways}determin.\end{sideways}}&\multirow{2}{*}{\begin{sideways}TED\end{sideways}}&validation&22.39&+0.59&+1.50&+1.89&25.43\\
		&&test&22.76&+0.67&+1.41&+2.02&25.58\\
		\cmidrule{2-8}
		&\multirow{2}{*}{\begin{sideways}News\end{sideways}}&validation&24.64&+0.62&+0.99&+1.02&27.62\\
		&&test&25.27&+0.94&+1.05&+1.13&28.08\\
		\midrule
		\midrule
		&&&out-of-domain&IPS+R&DR&\cDR{}&in-domain\\
		\midrule
		\multirow{4}{*}{\begin{sideways}\hskip -0.3cm stochastic\end{sideways}}&\multirow{2}{*}{\begin{sideways}TED\end{sideways}}&validation&22.39&+0.57&+1.92&+1.95&25.43\\
		&&test&22.76&+0.58&+2.04&+2.09&25.58\\
		\cmidrule{2-8}
		&\multirow{2}{*}{\begin{sideways}News\end{sideways}}&validation&24.64&+0.71&+1.00&+0.71&27.62\\
		&&test&25.27&+0.81&+1.18&+0.95&28.08\\
		\bottomrule
	\end{tabular}
\end{table*}

As shown in Table \ref{exp:learn_results}, under deterministic logging, the best results are obtained by the combining reweighting and double control in the \cDC{} method. The relations between the algorithms and even the absolute improvements are quite similar under stochastic logging. For an extended discussion see \cite{LawrenceETAL:17}.

	\section{Discussion}

We presented an analysis of possible degenerate behavior in counterfactual learning scenarios. We analyzed the degeneracies of the standard inverse propensity scoring method and its weighted variant, both under stochastic and deterministic logging. Our analysis facilitates a clearer understanding why doubly robust learning techniques serve to avoid such degeneracies, and why such techniques even allow to perform counterfactual learning under deterministic logging.

\cite{LawrenceETAL:17} also discuss a possible implicit exploration effect by the stochastic selection of inputs. This phenomenon has recently been given a formal account by \cite{BastaniETAL:17} and has not been analyzed formally in this paper.

An open question is the application of the techniques proposed by \cite{LawrenceETAL:17} to machine translation with neural networks. For example, the necessity to normalize probabilities over the full set of logged data creates a memory bottleneck which makes it difficult to transfer the reweighting approach to neural networks.

	\subsubsection*{Acknowledgments}
	
	The research reported in this paper was supported in part by the German research foundation (DFG).

	\bibliography{ref}
	\bibliographystyle{apalike}
	
\end{document}